\documentclass[submission,copyright,creativecommons]{eptcs}
\providecommand{\event}{FMAS 2025} % Name of the event you are submitting to
\usepackage{iftex}
\usepackage{verbatim}
\usepackage{graphicx}
\usepackage{subcaption}
\usepackage[final]{fixme}
\usepackage{amsfonts,amsmath,stmaryrd}
\usepackage{wrapfig}
\usepackage{soul}
\usepackage{xcolor}
\sethlcolor{yellow} 

\newcommand{\DP}{\mathit{DP}}
\newcommand{\NN}{\mathit{NN}}
\newcommand{\Class}{\mathrm{Class}}
\newcommand{\target}{\mathrm{target}}
\newcommand{\Power}{\mathrm{Power}}

\newcommand{\edit}[1]{{\color{black} #1}}

\ifpdf
  \usepackage{underscore}         % Only needed if you use pdflatex.
  \usepackage[T1]{fontenc}        % Recommended with pdflatex
\else
  \usepackage{breakurl}           % Not needed if you use pdflatex only.
\fi

\title{Formal Verification of Local Robustness of a Classification Algorithm for a Spatial Use Case%
\thanks{This work was supported by the EU Horizon Research and Innovation Actions program under Grant 101070162.}}

\author{
Delphine Longuet \qquad Amira Elouazzani
\institute{Thales, TRT, cortAIx Labs\\ France}
\email{name.surname@thalesgroup.com}
\and
Alejandro Penacho Riveiros \qquad Nicola Bastianello
\institute{KTH Royal Institute of Technology\\ Stockholm, Sweden}
\email{\{alejpr|nicolba\}@kth.se}
}

\def\titlerunning{Formal Verification of Local Robustness of a Classification Algorithm for a Spatial Use Case}
\def\authorrunning{D. Longuet, A. Elouazzani, A. Penacho Riveiros \& N. Bastianello}
\def\copyrightholders{Longuet, Elouazzani, Penacho Riveiros \& Bastianello}

\begin{document}
\maketitle

\begin{abstract}
Failures in satellite components are costly and challenging to address, often requiring significant human and material resources. Embedding a hybrid AI-based system for fault detection directly in the satellite can greatly reduce this burden by allowing earlier detection. However, such systems must operate with extremely high reliability. To ensure this level of dependability, we employ the formal verification tool Marabou to verify the local robustness of the neural network models used in the AI-based algorithm. This tool allows us to quantify how much a model’s input can be perturbed before its output behavior becomes unstable, thereby improving trustworthiness with respect to its performance under uncertainty.
\end{abstract}

%\section{Introduction}
The number of satellites orbiting Earth and their importance have significantly increased in recent decades~\cite{McDowell2020}. Indeed, they serve as the infrastructure for fundamental technologies such as positioning and communication, and enable scientific advancements in Earth observation.
However, these satellites are deployed in a difficult environment, which might lead to failures, and are in practice inaccessible for maintenance. Therefore, there is a need to ensure these satellites function without failures throughout their mission.
Among the components prone to failure are the Reaction Wheel Assemblies (RWAs), which consist of an electric motor rotating a disk that imparts momentum to the satellite to change its attitude~\cite{narkiewicz2020generic}. Thus RWAs have mechanical components in constant motion, making them prone to failures. Given the cost of a failure (potentially terminating a mission), it is important to detect and alert the ground station of possible impending failures.
In this paper thus we focus on a hybrid approach, combining data-driven technique and physics-based insights, to detect anomalies in RWAs that might indicate a future failure. In particular, we focus on detecting four types of anomalies, related to different components of the RWA, each of which has three urgency levels. The full hybrid algorithm is described in~\cite{riveiros2024realtime,riveiros2025hybrid}.
Given the importance of preventing failures in RWAs, the anomaly detection and classification algorithm should be reliable, avoiding false negatives especially, but also false positives to avoid burdening the ground station personnel. Additionally, it should be robust to perturbations in the data, due to \textit{e.g.} unforeseen operating conditions or degradation of the components. Therefore, in this paper we focus on guaranteeing the local robustness of the hybrid algorithm through formal verification, targeting in particular the neural networks that are part of the classification algorithm.

An AI classification model is said to be locally robust\footnote{Local robustness is also called stability, for example in the new ML standard ED-324/ARP6983 (ongoing work from the joint EUROCAE/SAE working groups WG-114/G-34).} for a given perturbation (noise) of the inputs if every input $x$ classified as $y$, all the inputs resulting from this perturbation applied to $x$ are also classified as $y$. It means that the model does not change the classification for an input if it is only slightly modified. 

In recent years, several formal verification tools have been developed to provide mathematical guarantees for AI algorithms, with a particular focus on neural networks~\cite{shi2024genbab,Pyrat24,nnenum21,Marabou24}. These tools take as input a trained neural network $N$ seen as a function $f : I \rightarrow O$ between a set of inputs $I$ and a set of outputs $O$, and an input-output relationship property of the form $\forall x \in I, ( P(x) \land f(x) = y ) \Rightarrow Q(y)$, where $P$ and $Q$ are predicates for pre- and post-conditions.
Based on these inputs, the tools use SMT-based (Satisfiability Modulo Theories) and abstract interpretation techniques to prove the property or find a counter-example. 

The classification algorithm developed for the detection of RWAs anomalies implements a hybrid approach where neural networks are used after an analytical data processing phase. Unfortunately, this algorithm was not designed in order to be formally verified, therefore applying formal verification to the whole algorithm is not currently possible. Instead, we propose a methodology that relies on formal methods to increase confidence in its local robustness, establishing local robustness of the neural networks that are part of the classification algorithm. \edit{Formal verification was performed with the tool Marabou based on SMT constraint-solving: it is particularly appropriate in our case since we deal with small-sized neural networks with ReLU activation for which its resolution is complete, and it offers a user-friendly Python interface.}

This paper is structured as follows. In Section~\ref{sec:algo}, we discuss the design of the hybrid algorithm~\cite{riveiros2024realtime,riveiros2025hybrid}, which combines data-driven tools and physics-based insights to detect and classify anomalies in satellites' RWAs. We summarize its structure and showcase some illustrative results in terms of classification accuracy. We highlight its resilience
on neural network components, motivating the need for a robustness analysis.
Section~\ref{sec:methodo} proposes a methodology to achieve local robustness verification of this classification algorithm, using formal verification of the neural networks involved. We explain how this methodology improves confidence even though formal verification of the whole algorithm is not tractable. Results are presented in Section~\ref{sec:results}.
In Section~\ref{sec:global}, we discuss how we go beyond local robustness by defining sets of inputs for which the classification is certain, thus improving confidence in the global robustness of the algorithm on some particular classes.

\edit{Although the considered data and classification algorithm are proprietary and confidential, we believe that the presented methodology and findings are of interest for other implementations of similar (and possibly other) AI algorithms involving neural networks, and can be reproduced thanks to the details we provide.}

\section{Algorithm Description}
\label{sec:algo}

In this section we briefly discuss the proposed anomaly classification algorithm, summarizing its design and showcasing its classification performance. The complete details can be found in~\cite{riveiros2024realtime,riveiros2025hybrid}.

\subsection{Problem Formulation}
\begin{figure}
    \begin{minipage}[t]{0.48\textwidth}
        \centering
        \includegraphics[width=\linewidth]{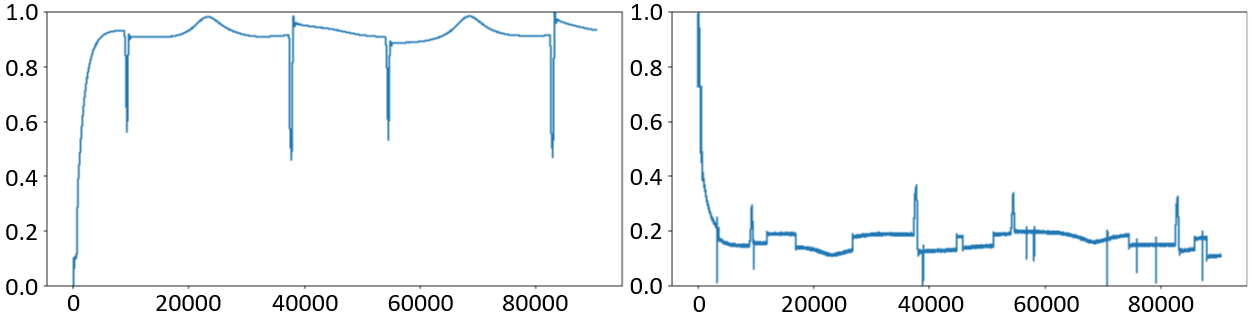}
        \caption{Time series of a RWA spin rate and friction torque}
        \label{fig:time-series}
    \end{minipage}
    \hfill
    \begin{minipage}[t]{0.5\textwidth}
        \centering
        \includegraphics[width=\linewidth]{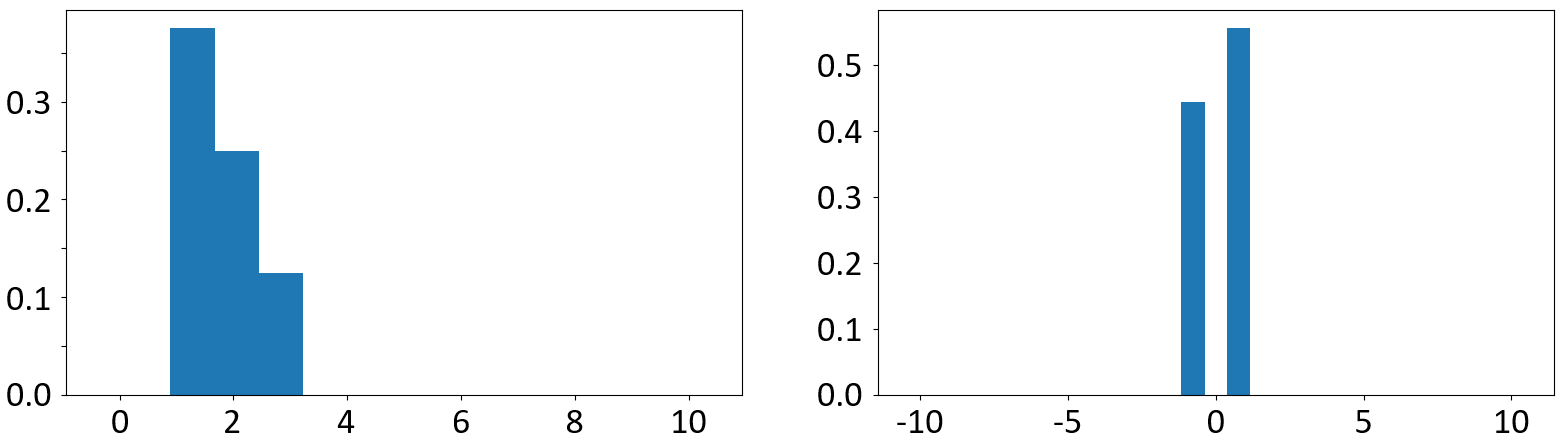}
        \caption{Histograms used to classify anomalies C and D}
        \label{fig:histograms}
    \end{minipage}    
%    \caption{time series of RWA spin rate time series (a) and corresponding histograms of coefficients C and D (b).}
%    \label{fig:series-histograms}
\end{figure}

We start by formally stating the problem we aim to solve. The goal is to use on-board measurements to classify the status of the RWA, which can be: nominal, or anomaly A, B, C, D, each of which has $3$ urgency levels.\footnote{\edit{Note that the four anomalies are anonymized for confidentiality reasons.}} A higher urgency level corresponds to a higher likelihood that the RWA will break down in the near future, and thus the ground station needs to be notified more quickly of the RWA's anomalous behavior.
The data available for the classification are measurements of the spin rate $\omega_k$ [rad/s] indexed by $k \in \mathbb{N}$, and measurements of the total friction torque $f_k$ [mNm]. One datapoint is then characterized as the sequence of measurements $\bigl( (\omega_k, f_k) \bigr)_{k = 1}^N$, $N \in \mathbb{N}$. Figure~\ref{fig:time-series} depicts an example of the spin rate and friction torque time series.
We remark that the friction torque is not directly measured, but rather estimated from the total output torque of the motor.

\textbf{Problem}: using spin rate and friction torque data, classify the status of the RWA as either nominal or anomalous, and in the latter case classify the type of anomaly (A, B, C, D) and its urgency level (1--3).

\subsection{Hybrid Algorithm Design}
In this section we summarize the design of the hybrid classification algorithm, which combines data-driven techniques with physics-based insights. We start by discussing the physical foundation of the algorithm.

\subsubsection{Physics Model}
The RWA is a rotating machine, and thus is a friction system characterized by a relationship between spin rate and the friction torque. In our design, we choose to employ the following simple model of this relationship~\cite{armstrong1991control}:
\begin{equation}\label{eq:friction-model}
    f_k = f_k^\mathrm{d} \operatorname{sign}(\omega_k) + f^\mathrm{v} \omega_k + v_k
\end{equation}
where:
\begin{itemize}
    \item $f_k^\mathrm{d} \operatorname{sign}(\omega_k)$ is the dry friction torque, with magnitude ($f_k^\mathrm{d}$) independent of the spin rate and always opposing the motion of the wheel. $f_k^\mathrm{d}$ is indexed by $k$ as it can change over time. Dry friction is caused by the contact between solid surfaces inside the bearings~\cite{longato2023microvibration}. Anomalies A, C, D affect $f_k^\mathrm{d}$.

    \item $f^\mathrm{v} \omega_k$ is the viscous friction torque, proportional to the spin rate of the wheel and with constant coefficient $f^\mathrm{v}$. This component is due to the lubricant present in the bearings of the RWA~\cite{lee2015flight}. We remark that, although the viscous friction coefficient can be affected by temperature and pressure~\cite{seeton2006viscosity}, these changes take place on a longer time scale than the anomalous phenomena we seek to classify in this paper. Anomaly B affects $f^\mathrm{v}$.
    
    \item $v_k$ is the noise in the system, which we assume to be normal with zero mean and variance $\sigma$, as motivated by \textit{e.g.}~\cite{carrara2013estimating}.
\end{itemize}

\edit{We remark that this model is not a precise representation of the RWA behavior, which is significantly more complex. Nonetheless, this model serves as a good foundation for the design of the classification algorithm when combined with labeled data.}

\smallskip

We turn now to describing the overall classification algorithm. The algorithm is composed of two parts: 1) data processing (see section~\ref{subsec:data-processing}) and 2) classification based on the processed data (see section~\ref{subsec:classification}).

\subsubsection{Data Processing}\label{subsec:data-processing}
The role of the data processing step is to extract, from the measurements $\bigl( (\omega_k, f_k) \bigr)_{k = 1}^N$, the information necessary to classify the anomalies. In particular, the data processing consists of the following steps, depicted also in Figure~\ref{fig:data-processing}:
\begin{figure}[!ht]
    \centering
    \includegraphics[width=0.75\linewidth,  trim={2cm 7.25cm 1cm 7cm}, clip]{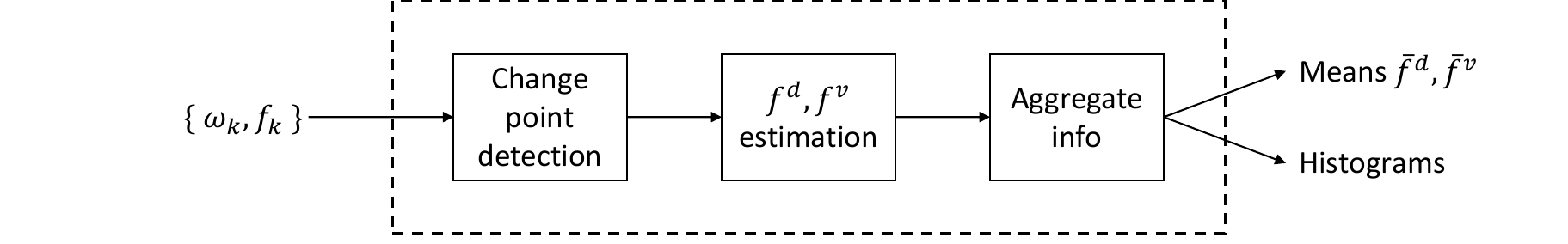}
    \caption{Data processing pipeline.}
    \label{fig:data-processing}
\end{figure}
\begin{enumerate}
    \item Friction coefficient estimation: the classification step requires the estimation of the values $f_k^\mathrm{d}$ and $f^\mathrm{v}$ from the sequence of measurements  $f_k$. Since model~\eqref{eq:friction-model} is linear, this can be easily done using linear regression. However, $f_k^\mathrm{d}$ undergoes instantaneous changes (both with nominal and anomalous status), and thus is better modeled as a piece-wise constant sequence of values. Thus the following step is required before the coefficients can be estimated. 

    \item Changepoint detection: since $f_k^\mathrm{d}$ is a piece-wise constant sequence of values, the idea is to break down the sequence of measurements in the intervals when $f_k^\mathrm{d}$ is constant. To this end, we need to detect when a change in its values occurs. This is performed by fitting the linear model~\eqref{eq:friction-model} against the sequence of measurements on a rolling window of size much smaller than $N$, \edit{which is a tunable hyperparameter -- see discussion in~\cite{riveiros2025hybrid} on how to choose it}. When the linear fit loses accuracy, it means that a change has occurred, and we can divide the sequence in two intervals where $f_k^\mathrm{d}$ has two different values. We continue this operation for the whole sequence, ending up with a set of intervals, to which we can individually apply friction coefficient estimation described in 1. We denote by $\hat{f}_i^\mathrm{d}$ and $\hat{f}_i^\mathrm{v}$ the estimated friction coefficients over the $i$-th interval.

    \item Coefficients estimates processing: after running steps 1. and 2., we have a set of intervals $[k_i, k_{i+1})$, $k_i \in \{ 1, \ldots, N \}$, $k_i < k_{i+1}$, indexed by $i \in \{ 1, \ldots, I \}$, and to each interval we associate the coefficients estimates $\hat{f}_i^\mathrm{d}$ and $\hat{f}_i^\mathrm{v}$. The last step then is to process the coefficients estimates to extract the information required for classification of the different anomalies. In particular, we extract the following:
    \begin{enumerate}
        \item The mean values of dry and viscous friction coefficients $\bar{f}^\mathrm{d} = \frac{1}{I} \sum_i \hat{f}_i^\mathrm{d}$ and $\bar{f}^\mathrm{v} = \frac{1}{I} \sum_i \hat{f}_i^\mathrm{v}$ -- these are used to classify anomalies A and B.

        \item The specific distribution of the values $\hat{f}_i^\mathrm{d}$ acts as a signature of the anomalies C and D. Therefore, we approximate the two anomalies' distributions via two histograms of the variations in friction coefficient, \textit{i.e.} $\hat{f}^\mathrm{d}_{i+1} - \hat{f}^\mathrm{d}_{i}$. In particular, the histogram for anomaly D collects the values of $\hat{f}^\mathrm{d}_{i+1} - \hat{f}^\mathrm{d}_{i}$ in $M$ bins, computing their relative frequency to build the histogram. On the other hand, anomaly C is characterized by an increase and subsequent decrease of roughly equal magnitude -- that is, $\hat{f}^\mathrm{d}_{i+1} - \hat{f}^\mathrm{d}_{i} \simeq - (\hat{f}^\mathrm{d}_{i+2} - \hat{f}^\mathrm{d}_{i+1})$. The histogram thus only needs to include information on the first variation in friction torque (the increase), and the $M$ bins collect only these values and their frequency. The construction of these histograms then proceeds as follows: we check for matching pairs of friction torque increase and decrease, and collect the magnitude of the increases in the anomaly C bin, discarding the decreases. The anomaly D histogram is then constructed with the magnitudes of the variations not accounted for before. Figure~\ref{fig:histograms} shows examples of anomaly C and D histograms.
    \end{enumerate} 
\end{enumerate}

\subsubsection{Classification}\label{subsec:classification}
\begin{figure}[!ht]
    \centering
    \includegraphics[width=0.75\linewidth,  trim={0 3.6cm 2cm 7.8cm}, clip]{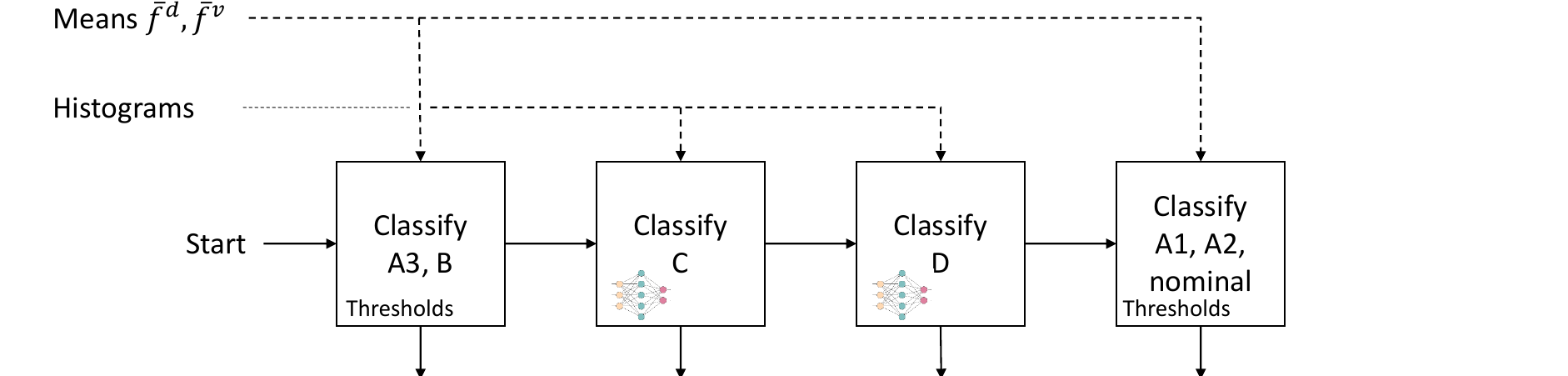}
    \caption{Classification algorithm overview.}
    \label{fig:classification}
\end{figure}

With the results of the data processing in place, we can now provide an overview of the classification algorithm. The algorithm is structured similarly to a decision tree, where different nodes classify different types of anomalies or whether the status is nominal, see Figure~\ref{fig:classification}.
The four classification modules are designed as follows:
\begin{itemize}
    \item Classification of A3, B: the classification of anomalies A, B and of nominal behavior can be done via the mean values $\bar{f}^\mathrm{d}$ and $\bar{f}^\mathrm{v}$. Indeed, anomaly A corresponds to an increase in $\bar{f}^\mathrm{d}$ as compared to the nominal baseline, while anomaly B corresponds to an increase in $\bar{f}^\mathrm{v}$. Additionally, the magnitude of this increase defines the urgency level of the anomaly. The first module then applies threshold checking to discriminate between A3 and B (B1, B2, B3). At this stage, we do not classify A1, A2, and nominal, as this might lead to the misclassification of anomalies C and D.

    \item Classification of C and D: the second and third modules then employ the histograms (approximating the distribution of the changes in $f_k^\mathrm{d}$) to classify C and D, and assign an urgency level as well. This classification is performed employing two small-scale neural networks, each characterized by 1 hidden layer of 10 neurons, with ReLU activation.

    \item Classification of A1, A2, nominal: finally, sequences that have not been classified by any of the previous modules are classified based on the values of $\bar{f}^\mathrm{d}$ and $\bar{f}^\mathrm{v}$, again using threshold checking.
\end{itemize}
The overall structure of the algorithm then is designed to minimize misclassification between the types of anomalies. The thresholds and neural networks are trained using examples of $\bigl( (\omega_k, f_k) \bigr)_{k = 1}^N$ labeled with an anomaly (or nominal) status, and with an urgency level. For training, we use $40\%$ of the available labeled data, which includes 2000 nominal, 683 anomaly A, 871 B, 792 C, 875 D, each divided equally by urgency level.

\subsection{Classification Results}
\label{subsec:classif-results}

In this section we briefly report the results on the accuracy of the classification algorithm applied to a validation set of data.
We start by presenting the confusion matrix of the classifier in Table~\ref{tab:confusion-matrix}, where the columns correspond to the true status and the rows correspond to status output by the algorithms.
\begin{table}[!ht]
    \centering
    \caption{Confusion matrix between actual status (columns) and status classified by the algorithm (rows). `N' stands for nominal, and an empty cell implies no confusion.}
    \renewcommand{\arraystretch}{1.5}
    \begin{footnotesize}
        \begin{tabular}{c|c|c|c|c|c|c|c|c|c|c|c|c|c|}
        \cline{2-14}
        D3 & 0.2 &  &  &  &  &  &  &  &  &  &  & 19.2 & 85.6 \\ \cline{2-14}
        D2 & 0.2 &  &  &  &  &  &  &  &  &  & 4.4 & 79.7 & 14.4 \\ \cline{2-14}
        D1 & 0.6 &  & 1.3 &  &  &  &  &  &  &  & 95.6 & 1.1 &  \\ \cline{2-14}
        C3 &  &  &  &  &  &  &  &  & 2 & 100 &  &  &  \\ \cline{2-14}
        C2 &  &  &  &  &  &  &  &  & 98 &  &  &  &  \\ \cline{2-14}
        C1 &  &  &  &  &  &  &  & 100 &  &  &  &  &  \\ \cline{2-14}
        B3 &  &  &  &  &  &  & 100 &  &  &  &  &  &  \\ \cline{2-14}
        B2 &  &  &  &  &  & 99.4 &  &  &  &  &  &  & \\ \cline{2-14}
        B1 &  &  &  &  & 99.4 & 0.6 &  &  &  &  &  &  &  \\ \cline{2-14}
        A3 &  &  &  & 99.4 &  &  &  &  &  &  &  &  &  \\ \cline{2-14}
        A2 &  & 1.1 & 95.6 & 0.6 &  &  &  &  &  &  &  &  &  \\ \cline{2-14}
        A1 & 0.9 & 92.1 & 3.1 &  &  &  &  &  &  &  &  &  &  \\ \cline{2-14}
        N & 98.1 & 6.7 &  &  & 0.6 &  &  &  &  &  &  &  &  \\ \cline{2-14}
        \multicolumn{1}{c}{} 
        & \multicolumn{1}{c}{N} 
        & \multicolumn{1}{c}{A1} 
        & \multicolumn{1}{c}{A2} 
        & \multicolumn{1}{c}{A3} 
        & \multicolumn{1}{c}{B1} 
        & \multicolumn{1}{c}{B2} 
        & \multicolumn{1}{c}{B3} 
        & \multicolumn{1}{c}{C1} 
        & \multicolumn{1}{c}{C2}
        & \multicolumn{1}{c}{C3} 
        & \multicolumn{1}{c}{D1} 
        & \multicolumn{1}{c}{D2} 
        & \multicolumn{1}{c}{D3} \\
    \end{tabular}
    \end{footnotesize}
    \label{tab:confusion-matrix}
\end{table}
As we can see, the classification accuracy is generally high when discriminating between different anomalies and nominal. However, there is some misclassifications between nominal and low urgency A, and some misclassifications between different urgency levels of the same anomaly.

We present now the results in terms of additional performance indicators. To define them, we denote by $N(\text{X})$ the total number of datapoints with status X (A-D or nominal), $N_c(\text{X})$ as the number of datapoints with status X correctly classified as X, and $N_{\text{all}}(\text{X})$ as the number of all datapoints classified as status X. Then we define sensitivity as $\text{S}(\text{X}) = N_c(\text{X}) / N(\text{X})$ and positive predictive value as $\text{PPV}(\text{X}) = N_c(\text{X}) / N_{\text{all}}(\text{X})$.
\begin{table}
    \centering
    \caption{Sensitivity and positive predictive values for different sub-sets of anomalies.}
    \begin{small}
    \begin{tabular}{cc|cc}
        Metric & Result [\%] & Metric & Result [\%] \\
        \hline
        S(A+B+C+D) & 99.6 & PPV(A+B+C+D) & 97.1 \\
        S(A1+B1+C1+D1) & 97.4 & PPV(A1+B1+C1+D1) & 86.6 \\
        S(A2+B2+C2+D2) & 92.8 & PPV(A2+B2+C2+D2) & 90.9 \\
        S(A3+B3+C3+D3) & 96.3 & PPV(A3+B3+C3+D3) & 94.8 \\
        S(A) & 98.0 & PPV(A) & 98.0 \\
        S(B) & 99.8 & PPV(B) & 100 \\
        S(C) & 100 & PPV(C) & 100 \\
        S(D) & 100 & PPV(D) & 97.4 
    \end{tabular}
    \end{small}
    \label{tab:kpi}
\end{table}
Table~\ref{tab:kpi} collects the resulting sensitivity and PPV for the overall classification, and divided by urgency level, and anomaly status. As we can see, these values are very high, exceeding $90\%$ in all but one case.

\section{Formal Verification Methodology for Local Robustness}
\label{sec:methodo}

Our goal is now to formally prove the local robustness (also called stability) of this classification algorithm.
We based our work on a formal verification tool for neural networks called Marabou~\cite{Marabou24}. Marabou is based on SMT (Satisfiability Modulo Theories) verification techniques and can check arbitrary linear properties over neural networks. Its result is either a formal guarantee
that the network satisfies the property, a concrete input for which the property is violated (a counterexample), or inconclusive \edit{(in case of timeout or incomplete resolution due to non-linear activation functions)}. 
In our case, on small-sized neural networks \edit{(1 layer of 10 neurons)} with only ReLU activation functions, Marabou can achieve a complete resolution, therefore the result is always conclusive.
It supports feed-forward neural networks with many different linear, piecewise-linear, and non-linear activation functions.
Properties can be specified using a Python interface or in a language dedicated to the verification of neural networks called VNN-LIB.\footnote{\url{https://www.vnnlib.org/}}
These specifications are expressed as input-output relationship properties. In our case, properties are defined by constraints on the inputs, as intervals around each datapoint, and constraints on the outputs, which ensure that the score for a targeted class remains greater than that of any other class. For a neural network $f : \mathbb{R}^N \to \mathbb{R}^M$, we define the classification function $\Class : \mathbb{R}^N \to \llbracket 1,M \rrbracket$ as
$
\Class(x) := \arg\max_{i \in \llbracket 1, M \rrbracket} f(x)_i$
Given two vectors $a,b \in \mathbb{R}^N$, \edit{with $a = (a_i)_{i=1}^N$ and $b = (b_i)_{i=1}^N$ representing the lower and upper bounds of each input datapoint}, and a targeted class $\target \in \llbracket 1,M \rrbracket$, a typical property can be expressed as:
\begin{equation}\label{eq:query}
\forall x \in \mathbb{R}^N, (\forall i \in \llbracket 1,N \rrbracket, a_i \leq x_i \leq b_i) \Rightarrow \Class(x) = \target
\end{equation}
As described in previous section, the classification algorithm is composed of two phases: a data processing phase, followed by a classification phase using thresholds and neural networks.
The output of the data processing phase is composed of several elements, among which are two histograms, used to classify anomalies C and D, using two different neural networks.
Our formal verification will focus on proving the local robustness of these two neural networks.

\subsection{Problem Formalization}

For a given dataset input \edit{$s = \bigl( (\omega_k, f_k) \bigr)_{k=1}^N \in (\mathbb{R}^2)^N
$}, the data processing phase of the algorithm returns two histograms \edit{$(h^\mathrm{C}, h^\mathrm{D}) \in (\mathbb{R}^M)^2$, each represented as a vector of bin values of length $M$, } that are used to classify anomalies C and D, respectively\footnotemark.
\footnotetext{We will simplify the notations $\_^\text{C}$ and $\_^\text{D}$ to $\_$ when the anomaly type is clear from context or irrelevant.}

We denote by $\DP : (\mathbb{R}^2)^N \to (\mathbb{R}^M)^2$ the data processing function, such that $\DP(s) = (h^\text{C}, h^\text{D})$. Each histogram is then passed to a neural network\footnotemark[\value{footnote}] $\NN^\text{C}, \NN^\text{D} : \mathbb{R}^M \to \llbracket 0,3 \rrbracket$, which assigns a class label corresponding to four levels: the absence of the anomaly (label 0), or its presence at one of three urgency levels (labels 1 to 3). Following the notations from previous sections, we denote these classes by $\{\text{no C}, \text{C1}, \text{C2}, \text{C3}\}$ and $\{\text{no D}, \text{D1}, \text{D2}, \text{D3}\}$, where label 0 indicates no anomaly.

The complete classification algorithms for anomalies C and D, as defined in the previous paragraph, are denoted by\footnotemark[\value{footnote}]
$\Class^{\text{C}}, \Class^{\text{D}} : (\mathbb{R}^2)^N \to \llbracket 0,3 \rrbracket$,
and are defined as compositions of $\DP$ with the respective networks:
$\Class^{\text{C}} = \NN^{\text{C}} \circ \DP$
and
$\Class^{\text{D}} = \NN^{\text{D}} \circ \DP$.

We denote by $P_\varepsilon^t(s)$ the set of inputs resulting from a chosen perturbation $t$ applied to an input $s$ with strength $\varepsilon \in \mathbb{R}^+$. \edit{Intuitively, $\varepsilon$ controls the magnitude of the distortion added to $s$, so that larger values of $\varepsilon$ correspond to noisier inputs further from the original series}. The local robustness property can then be formulated as follows: the classification algorithm is locally robust for a time series $s$ under perturbation $t$ with strength $\varepsilon$ if and only if
\[
\forall x \in (\mathbb{R}^2)^N,\ x \in P_\varepsilon^t(s) \Rightarrow \Class(x) = \Class(s).
\]

In the following, we focus on a single neural network, as the same methodology is applied independently to both classifiers of anomaly C and D.

\subsection{Perturbations}

\begin{figure}[t]
\begin{subfigure}{.5\textwidth}
    \includegraphics[width=\textwidth]{figures/og-serie.png}
    \caption{original time series}
    \label{ogserie}
\end{subfigure}%
\begin{subfigure}{.5\textwidth}
\includegraphics[width=\textwidth]{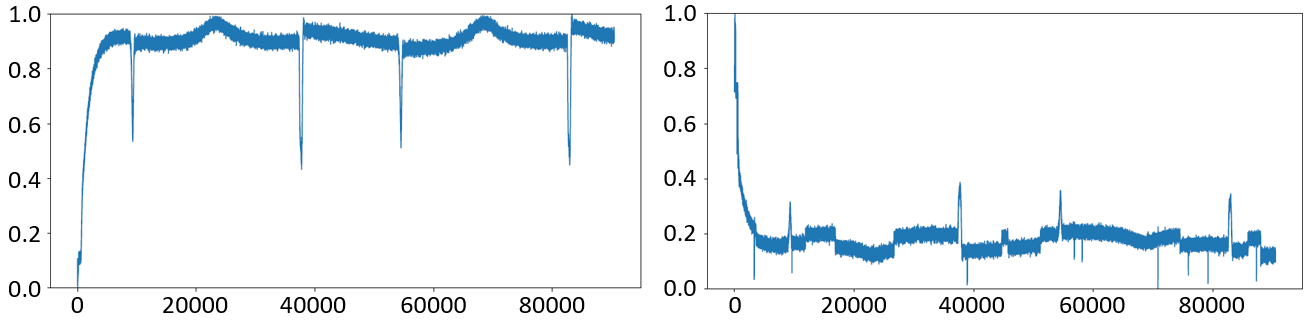}
    \caption{gaussian perturbation $\varepsilon = 0.01$}
    \label{gserie}
\end{subfigure}\\
\begin{subfigure}{.5\textwidth}
\includegraphics[width=\textwidth]{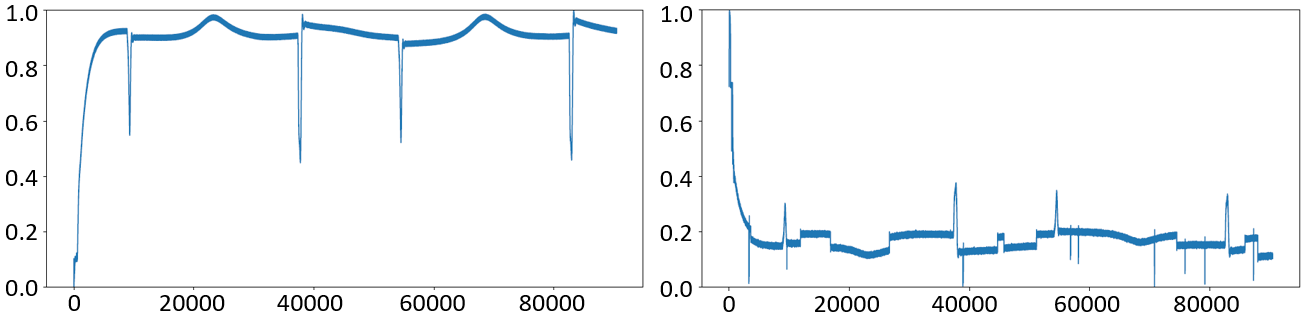}
    \caption{uniform perturbation $\varepsilon = 0.01$}
    \label{userie}
\end{subfigure}%
\begin{subfigure}{.5\textwidth}
\includegraphics[width=\textwidth]{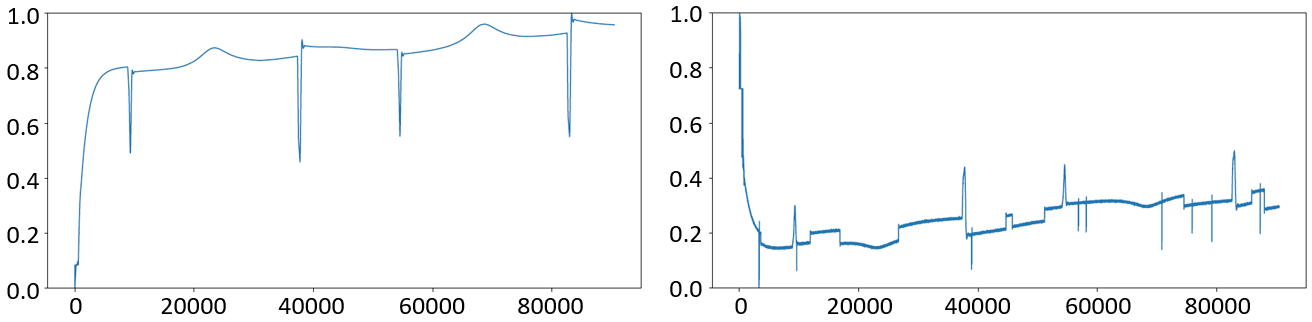}
    \caption{linear trend $\varepsilon = 0.2 $}
    \label{lserie}
\end{subfigure}
\caption{Example of time series and effect of some perturbations}
\label{perturbations}
\end{figure}

For classical perturbations modeled by an $L_\infty$ norm, for instance when inputs are images, we have $P_\varepsilon^t(s) = \{ x ~|~ \| s-x \|_\infty \leq \varepsilon \}$. In our case, we need to consider perturbations relevant to time series. Our discussions with domain experts led to the choice of the following perturbations: random noises (Gaussian, uniform, and Poisson), linear trend perturbation, missing data, and amplitude scaling.
\begin{itemize}
\item Random perturbations add to the original signal a noise, which is sampled from the chosen distribution then scaled by parameter $\varepsilon$ with respect to the signal amplitude (meaning that if $\varepsilon = 0.01$, the noise is scaled up to 1\% of the spin rate, resp. friction, amplitude). 
We chose Gaussian distribution with mean 0 and variance 1 (see Fig~\ref{gserie}), uniform distribution with values between $-1$ and 1 (see Fig~\ref{userie}), and Poisson distribution with parameter 1. 
\item The linear trend transformation simulates a sensor degradation which makes a linear bias with derivative $\varepsilon$ to be added to the series, causing the time series to slowly diverge, as can be seen in Fig~\ref{lserie}.
\item The amplitude scaling transformation simulates a change of magnitude of percentage $\varepsilon$ of the whole time series.
\item The missing data perturbation simulates a loss of data due to communication problems, for example, and corresponds to randomly removing a percentage $\varepsilon$ of datapoints.
\end{itemize}

\subsection{Methodology}

Formally verifying the local robustness of the whole classification algorithm for a given time series $s$ and perturbation $t$ at strength $\varepsilon$, as stated above, would boil down to formally computing the set $\DP(P_\varepsilon^t(s))$ and then verifying via Marabou that $\NN(\DP(P_\varepsilon^t(s))) \subseteq \Class(s)$. Unfortunately, this end-to-end formal verification process is not directly applicable to this classification algorithm. The formal computation of the set $\DP(P_\varepsilon^t(s))$ is not tractable by the tools we know: the data processing phase is developed in Python, a language for which no formal analysis tool exists, and consists of complex mathematical computations that are hardly eligible for formal reasoning. Furthermore, in order to use the tool Marabou, we need to express the property to verify with interval constraints on the datapoints of the input.

Therefore, we decided to approximate the computation of the set $\DP(P_\varepsilon^t(s))$ in the following way. For a given \edit{input} time series  $s$, we apply the chosen perturbation to $s$ a number of times \edit{determined experimentally}, to obtain a finite set of perturbed time series $S \subseteq P_\varepsilon^t(s)$. Then we compute $\DP(S)$, the result of the data processing phase on this set of inputs to obtain their relevant anomaly C or D histograms. Finally, we determine an envelope around these histograms that intuitively should represent the set of all histograms that we could obtain by applying this perturbation at this strength on $s$. \edit{We denote by $(u, v) \in (\mathbb{R}^M)^2$ the component-wise interval bound vectors of the histograms in $\DP(S)$, defined as
$\displaystyle
\forall k \in \llbracket 1, M \rrbracket, % \quad
u_k = \min_{h \in \DP(S)} h_k, %\quad
v_k = \max_{h \in \DP(S)} h_k,
$
which is equivalent to
$
\forall h \in \DP(S), \forall k \in \llbracket 1, M \rrbracket, %\quad
u_k \leq h_k \leq v_k.
$
}

Then we denote by $Env_\varepsilon^t(s)$ the envelope containing all histograms between the two vectors $u$ and $v$: $Env_\varepsilon^t(s) = \{ h \in \mathbb{R}^M ~|~ u_k \leq h_k \leq v_k \}$.
Note that this last set is neither an over- nor an under-approximation of the actual set $\DP(P_\varepsilon^t(s))$, since we may miss some histograms that fall outside of this envelope, and we may also include histograms that cannot be resulting from a perturbation of the initial input. However, we experimentally established that this envelope is a rather good approximation.

\begin{figure}
\begin{minipage}{0.49\textwidth}
    \includegraphics[width=\linewidth]
    {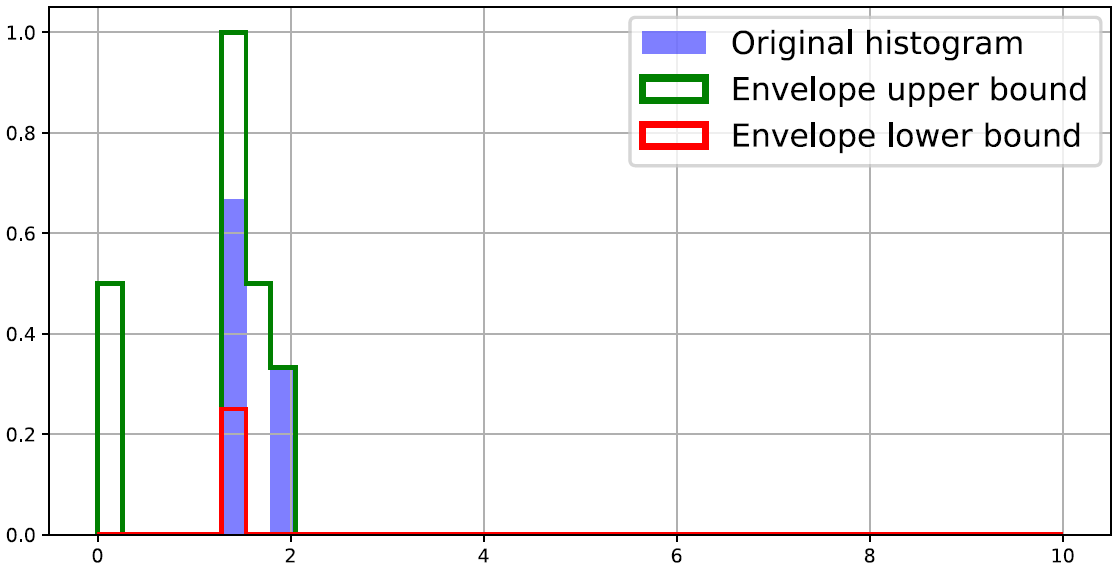}
    \caption{Envelope around a C1 histogram with a Gaussian noise at 0.005}
    \label{computation-envelope-histograms}
\end{minipage} \hfill
\begin{minipage}{0.49\textwidth}
    \includegraphics[width=\linewidth]
    {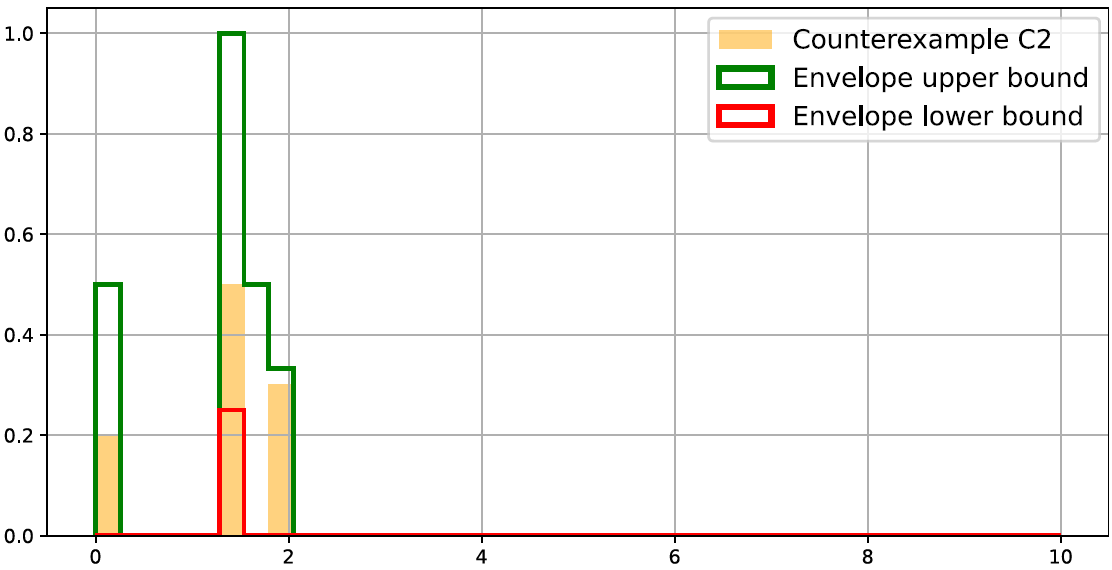}
    \caption{Counterexample found by Marabou in the envelope of Fig.~\ref{computation-envelope-histograms}}
    \label{counterexample}
\end{minipage}
\end{figure}

For example, we show on Figure~\ref{computation-envelope-histograms} the result of the computation of this envelope on a time series initially classified as an anomaly C1, whose histogram $h^\text{C}$ is shown in blue. We applied 10 times a Gaussian noise of strength $\varepsilon = 0.005$ (meaning 0.5\% of the amplitude of the signal) on this time series and obtained the envelope around their histograms determined by vectors $u$ (in green) and $v$ (in red).

%Figure~\ref{computation-envelope-histograms} shows in blue the initial anomaly C histogram $h^\text{C}$, an in green and red the vectors $u$ and $v$ respectively, determining the envelope around the histograms of 10 perturbations of $s$ obtained by applying a Gaussian noise at strength $\varepsilon = 0.006$ (meaning 0.6\% of the amplitude of the signal).

To choose the number of perturbation iterations for envelope construction, we tested several values and observed the evolution of the envelope. \edit{With the envelope width increasing by about 2\% when going from 10 to 100 iterations, while the computational cost increases by roughly 9 times, we fixed the number of iterations at 10 as a good compromise between accuracy and efficiency.
}

For the perturbations that are generated with a randomness factor (Gaussian, Poisson and uniform noises, as well as missing data perturbation), perturbing the same time series 10 times produces different series, allowing us to create 10 distinct histograms, even though at low strengths, some histograms may be similar due to data processing. For other perturbations like linear trend, the perturbation is deterministic, so we explicitly vary the strength across a fixed range to generate different versions. For amplitude scaling, which is also deterministic, we similarly apply a sequence of increasing strengths (e.g., from 0.5$\varepsilon$ to $\varepsilon$) to ensure we cover a range of effects and build a meaningful envelope.

Then we want to prove that all the histograms in this envelope are classified as the initial time series~$s$. The tool Marabou answers queries in the form of formula~(\ref{eq:query}) by transforming their negation into constraint satisfaction problems. It solves these problems by either providing an example that satisfies the problem (followed by a counterexample to the initial property) or proving that the problem is unsatisfiable (thus demonstrating the initial property). Then it has to solve 3 different queries, one for each $\target$ class different from $\Class(s)$, \edit{expressed by the following existential formula}:
$$\exists h \in \mathbb{R}^M, h \in Env_\varepsilon^t(s) \land \NN(h) = \target$$ 

\edit{Each of these queries leads to 13 constraints over 60 variables, which is well within Marabou's solving capacity.}
If one of these queries is satisfiable, the histogram found is a counterexample to local robustness. Otherwise, we have proof that the neural network is locally robust to this perturbation at this strength, for $s$. Figure~\ref{counterexample} shows, in orange, a histogram found by Marabou that is within the envelope computed around the C1 histogram of Figure~\ref{computation-envelope-histograms}, but it is classified as a C2 anomaly, thus demonstrating the non-robustness of the neural network around this histogram. 

\section{Evaluation Results for Local Robustness}
\label{sec:results}

We conducted our evaluation of local robustness only on correctly classified time series taken from series classified as anomaly C (urgency 1 to 3) and anomaly D (urgency 1 to 3), as well as two samples of series classified as `no C' and `no D'. The set of series classified as `no C' is sampled from the set of all the time series that pass through the neural network responsible for classifying anomaly C but are not classified as an anomaly C (see Fig.~\ref{fig:classification}). This includes all the time series that are classified as anomaly D (urgency 1 to 3), anomaly A urgency 1 and 2, and nominal. Likewise, the set of series classified as `no D' by the other neural network are all the series classified as anomaly A urgency 1 and 2, and nominal. 
To maintain coherence with the distribution of the data, we sampled 300 series for each of these classifications. For `no C', we sampled 50 entries from each of the 6 classes. For `no D' we sampled 100 entries from each of the 3 classes.  

To determine the relevant strengths to apply the chosen perturbations, we used the Signal-to-Noise Ratio (SNR) for each perturbation. For $s$ a time series defined on $n$ time units, $p$ a perturbed version of this series defined on the same time frame:
$$ \text{SNR}(s,p) = 10 \log_{10} \left(\frac{\Power(s)}{\Power(s-p)}\right) \text{\qquad where } \Power(s) = \frac{\sum_{k=1}^n s(i)^2}{n}$$

%\begin{wrapfigure}{r}{0.45\textwidth}
\begin{figure}
\begin{center}
\includegraphics[width=0.43\textwidth]{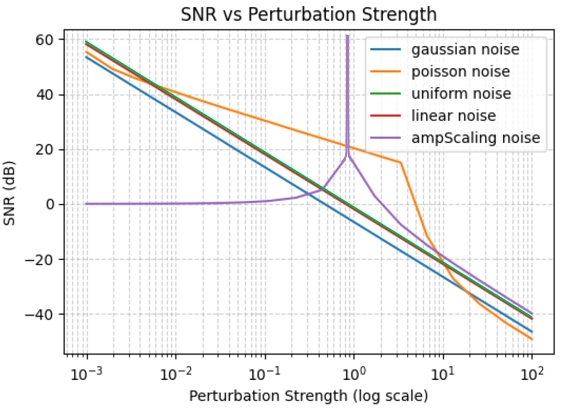}
\end{center}
\caption{Signal-to-Noise Ratio for each perturbation}
\label{SNR-by-perturbation}
\end{figure}
%\end{wrapfigure}

The SNR then quantifies the degradation of the signal: the higher the SNR, the more powerful is the actual signal, the lower the SNR, the more noise is present and the more information is lost. Figure~\ref{SNR-by-perturbation} displays the SNR for each perturbation, except missing data whose SNR cannot be computed due to the missing values. As our goal is to identify the boundary at which local robustness is lost, and following domain experts knowledge, we tested perturbations with SNRs down to 20~dB for amplitude scaling and 35~dB for all other perturbations, under which the classifier always fails to remain robust. 

Then we chose a sample of strength values from 0.001 up to the SNR threshold, on the logarithmic scale. 
The idea is to determine the maximal acceptable strengths required to keep the classification stable.

\begin{figure}[!ht]
\begin{subfigure}{.5\textwidth}
    \includegraphics[width=\textwidth]{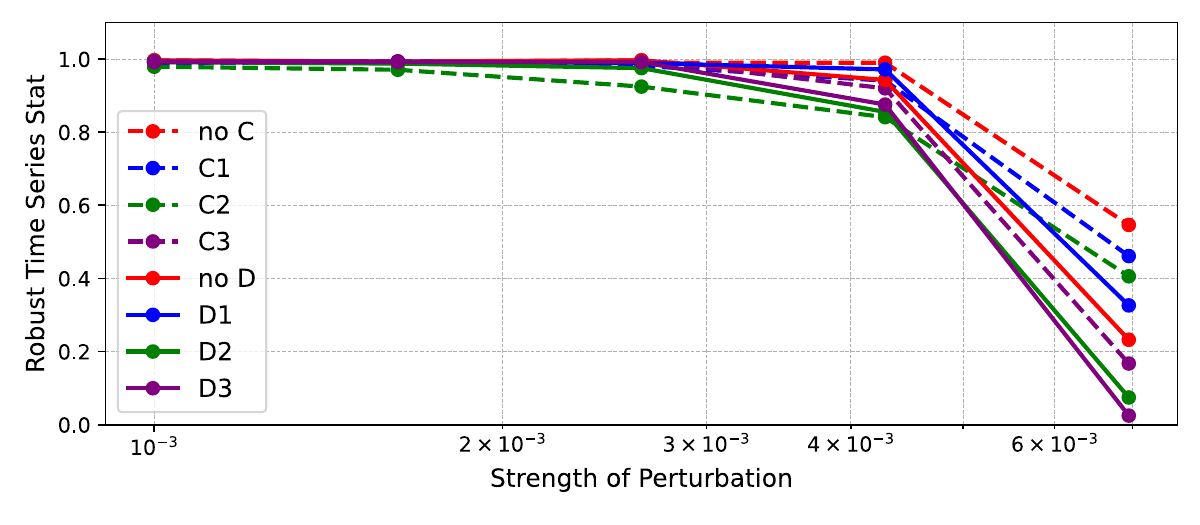}
    \caption{Gaussian noise}
    \label{gaussian}
\end{subfigure}%
\begin{subfigure}{.5\textwidth}
\includegraphics[width=\textwidth]{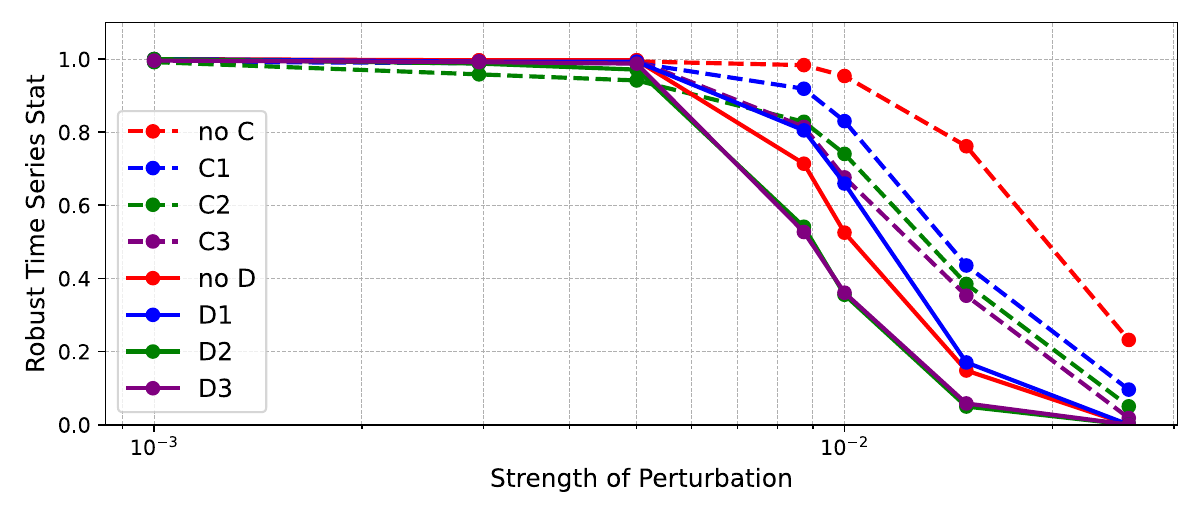}
    \caption{Poisson noise}
    \label{poisson}
\end{subfigure}\\
\begin{subfigure}{.5\textwidth}
\includegraphics[width=\textwidth]{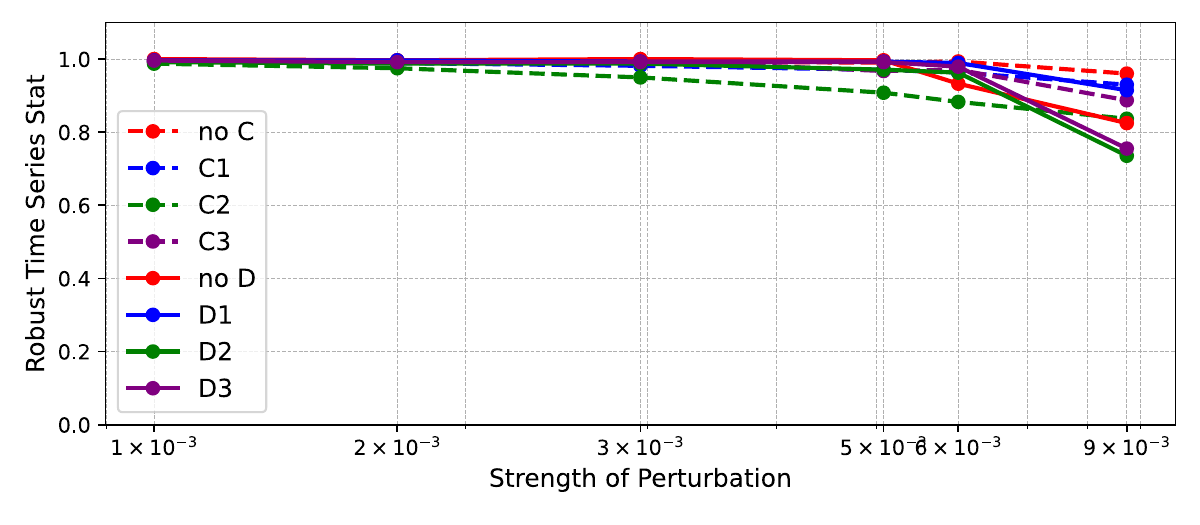}
    \caption{Uniform noise}
    \label{uniform}
\end{subfigure}%
\begin{subfigure}{.5\textwidth}
\includegraphics[width=\textwidth]{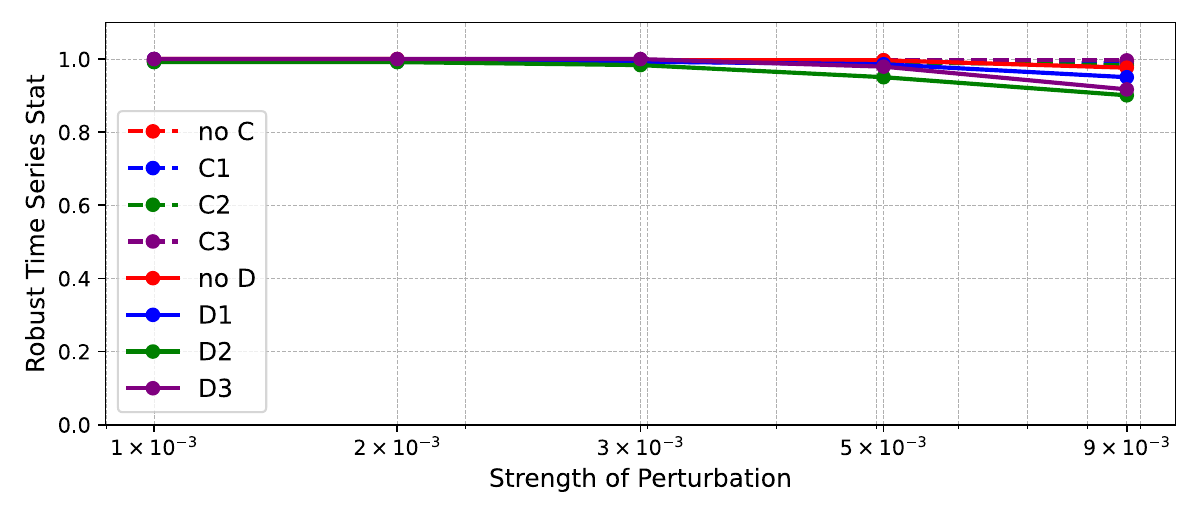}
    \caption{Linear trend}
    \label{linear}
\end{subfigure}\\
\begin{subfigure}{.5\textwidth}
\includegraphics[width=\textwidth]{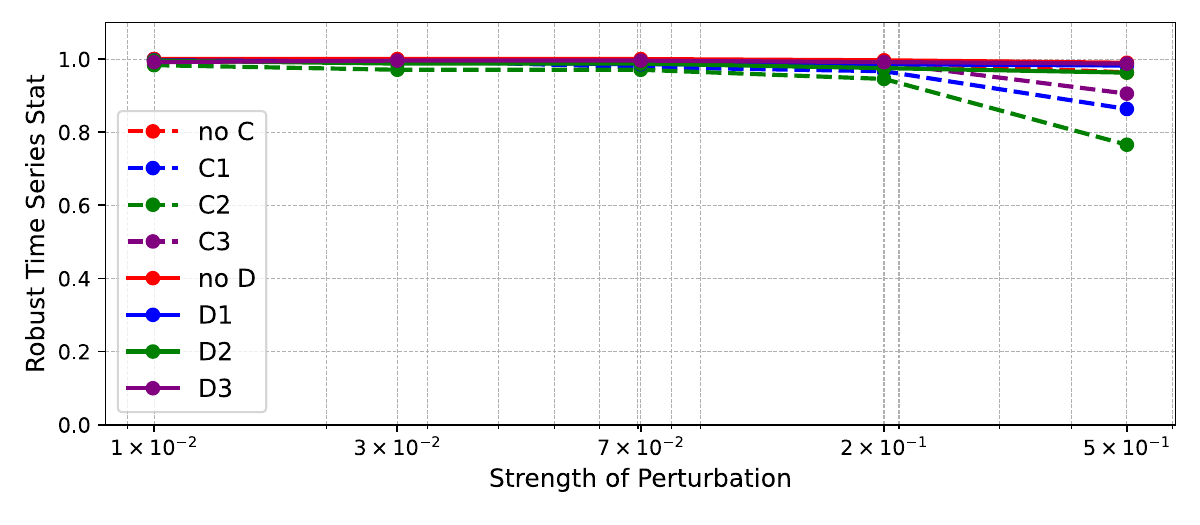}
    \caption{Missing data}
    \label{missing}
\end{subfigure}%
\begin{subfigure}{.5\textwidth}
\includegraphics[width=\textwidth]{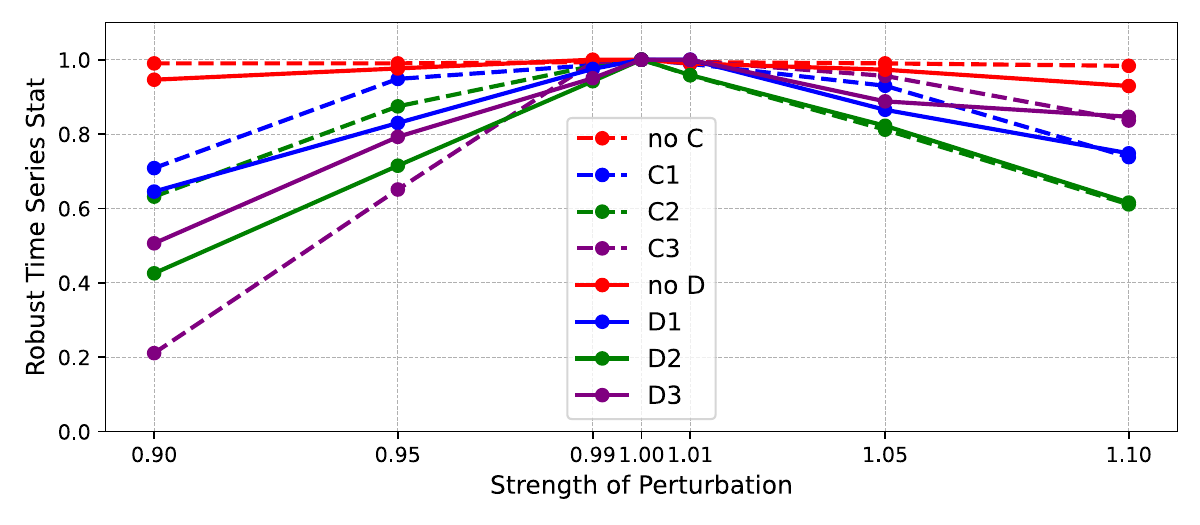}
    \caption{Amplitude shifting}
    \label{amplitude}
\end{subfigure}
\caption{Local robustness of neural networks classifying anomaly C and D to the different perturbations, class by class}
\label{robustness-results}
\end{figure}

In Figure~\ref{robustness-results}, we show the rate of time series for which the local robustness property is proven for each class and each strength of each perturbation.
For a given class $X$, as in section~\ref{subsec:classif-results}, we denote by $N_c(X)$ the number of time series with status $X$ correctly classified as $X$. For a perturbation $t$ and a strength $\varepsilon$, we also denote by $N_{lr}(X,t,\varepsilon)$ the number of such time series for which the entire envelope $Env^{\,t}_{\varepsilon}(s)$ is also classified as $X$. We then define the local robustness rate as $N_{lr}(X,t,\varepsilon) / N_c(X)$. This indicator therefore measures the fraction of correctly classified time series of class $X$ that remain robust under perturbation $t$ of intensity $\varepsilon$.

\begin{itemize}
\item Figure~\ref{gaussian},~\ref{poisson}, and~\ref{uniform}: we can see that for similar SNRs, the two neural networks are much more robust to uniform noise than to other random noises, Poisson noise being the most challenging for classification. 
\item Figure~\ref{linear}: linear trend perturbation for similar SNRs does not radically modify classification. 
\item Figure~\ref{missing}: we applied the missing data perturbation with different percentages of missing data, from 0.1\% to 50\% and did not notice a radical change in classification for most of the classes. 
\item Figure~\ref{amplitude}: we can see that the two neural networks are quite sensitive to amplitude scaling, more than a 2\% scaling strongly affects classification. 
\end{itemize}

Since our neural networks are primarily designed to detect the presence of anomalies, local robustness is particularly important to ensure that the network does not fail to detect an anomaly (false negative) or incorrectly classifies a nominal instance as anomalous (false positive). This means that if, during the verification process for an anomaly of type C (regardless of urgency), we find a counterexample classified as a non-anomaly, then there is a risk that our model would fail to detect the anomaly under perturbation. In order to focus on the robustness of anomaly detection
regardless of the anomaly urgency, we refined our previous robustness results by considering any misclassifications of an anomalous time series as nominal as a false positive.

\begin{figure}[t]
\begin{subfigure}{.5\textwidth}
    \includegraphics[width=\textwidth]{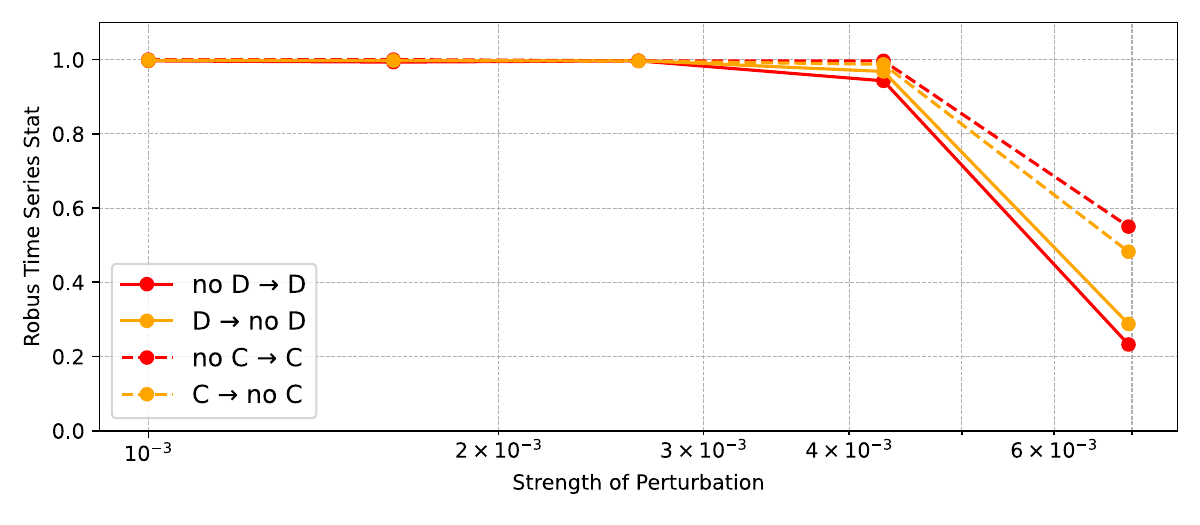}
    \caption{Gaussian noise}
    \label{gaussian anomaly no anomaly}
\end{subfigure}%
\begin{subfigure}{.5\textwidth}
\includegraphics[width=\textwidth]{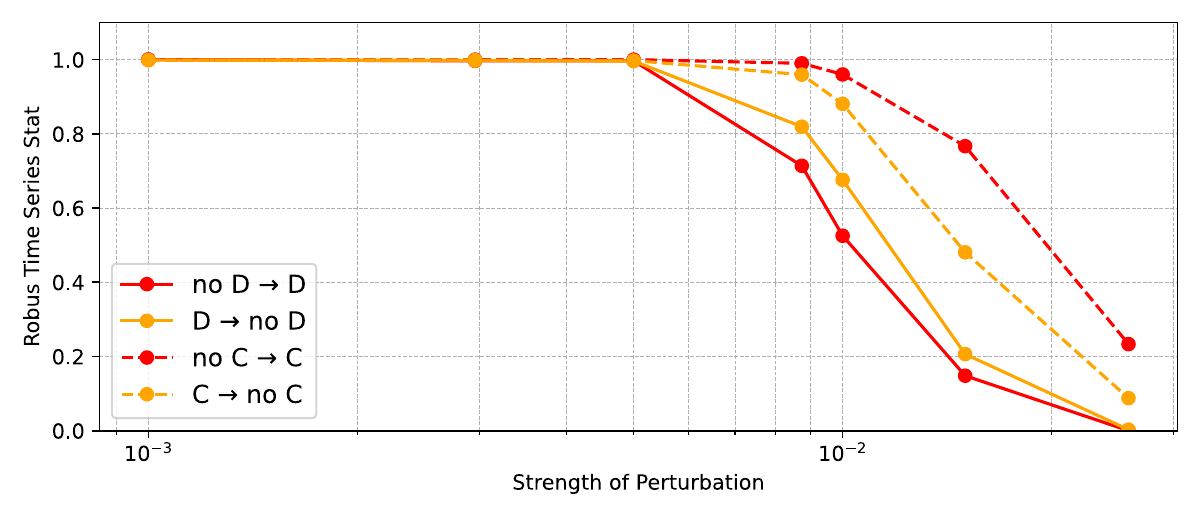}
    \caption{Poisson noise}
    \label{poisson anomaly no anomaly}
\end{subfigure}\\
\begin{subfigure}{.5\textwidth}
\includegraphics[width=\textwidth]{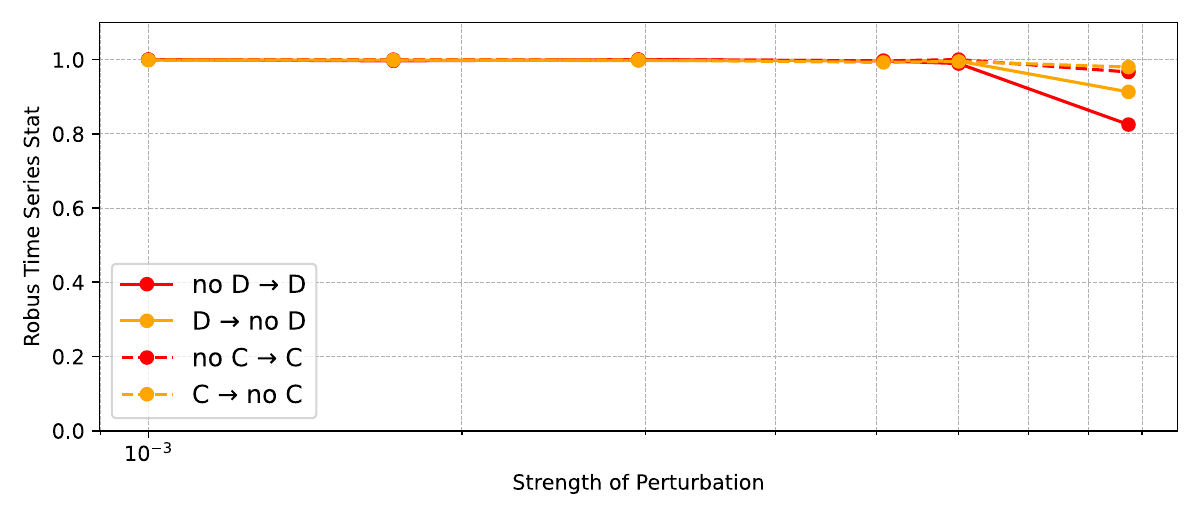}
    \caption{Uniform noise}
    \label{uniform anomaly no anomaly}
\end{subfigure}%
\begin{subfigure}{.5\textwidth}
\includegraphics[width=\textwidth]{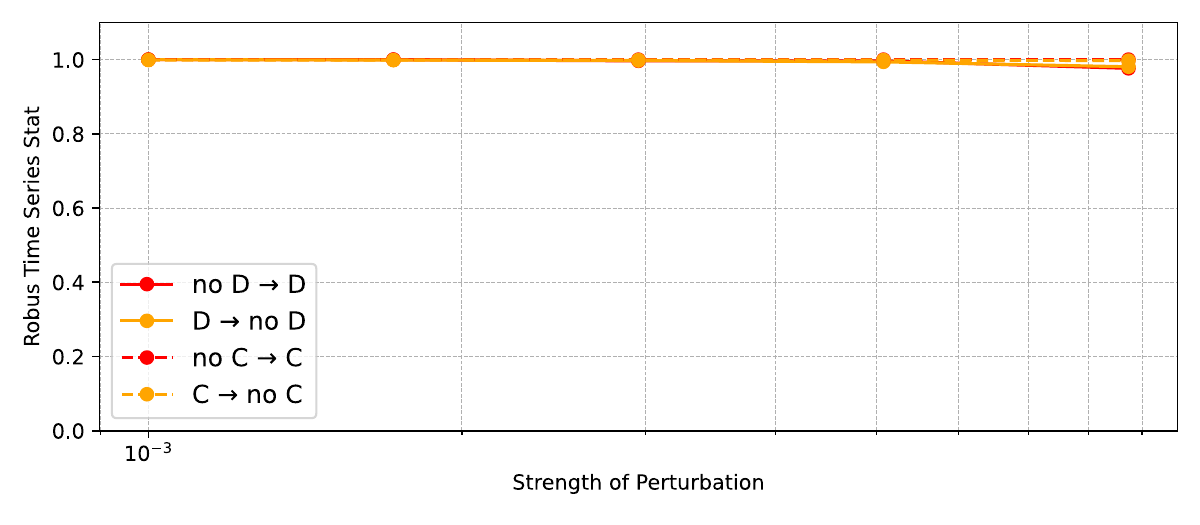}
    \caption{Linear trend}
    \label{linear anomaly no anomaly}
\end{subfigure}\\
\begin{subfigure}{.5\textwidth}
\includegraphics[width=\textwidth]{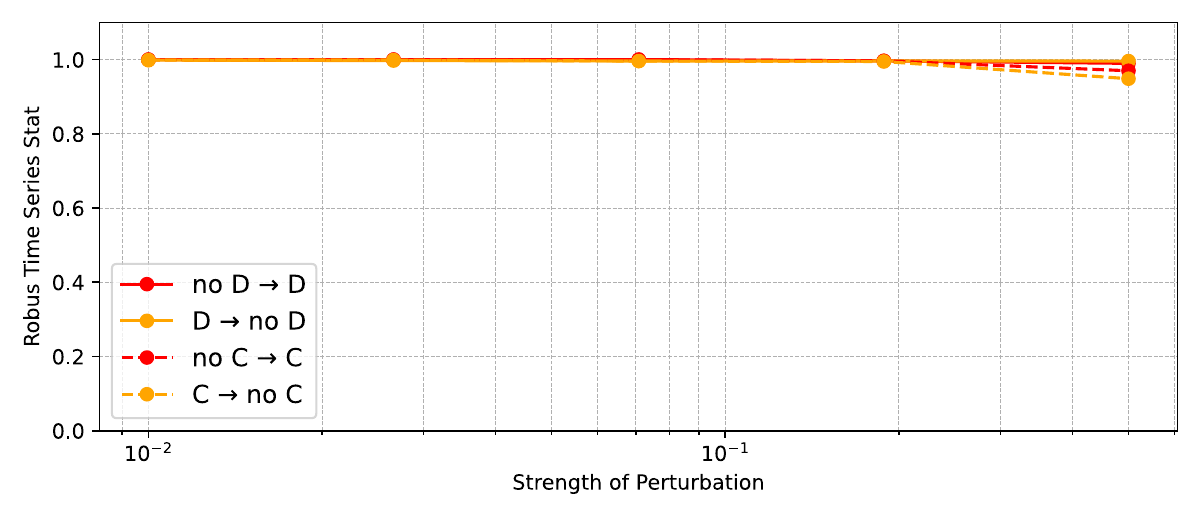}
    \caption{Missing data}
    \label{missing anomaly no anomaly}
\end{subfigure}%
\begin{subfigure}{.5\textwidth}
\includegraphics[width=\textwidth]{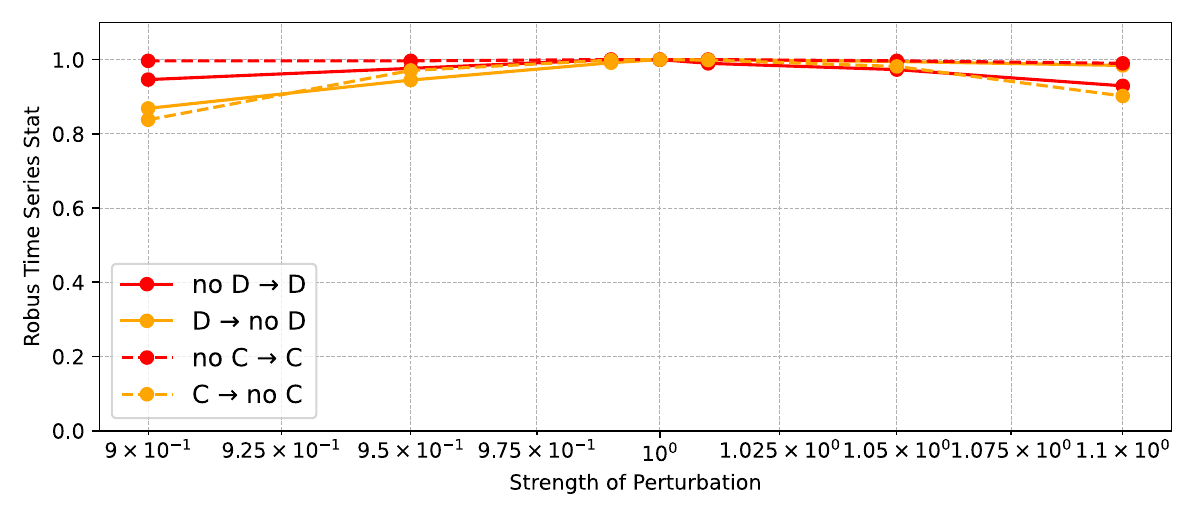}
    \caption{Amplitude shifting}
    \label{amplitude anomaly no anomaly}
\end{subfigure}
\caption{Local robustness of neural networks classifying anomaly C and D to the different perturbations, with respect to the detection of the anomaly}
\label{generalised-robustness-results}
\end{figure}

In Figure~\ref{generalised-robustness-results}, we visualize this property by distinguishing between the presence and absence of anomalies, as well as by the strength of each type of perturbation. Accordingly, we redefine 
$N_{lr}(X,t\varepsilon)$ by interpreting $X$ as a binary variable, distinguishing between C and `no C', as well as D and `no D'. We notice that, when considering only anomaly detection, all the results are better than those displayed in Fig.~\ref{robustness-results}, even if the general tendency remains the same for almost all perturbations.
\begin{itemize}
    \item Figures~\ref{gaussian anomaly no anomaly},~\ref{poisson anomaly no anomaly}, and~\ref{uniform anomaly no anomaly}: For similar SNRs, Gaussian and Poisson noise show a higher likelihood of the model missing anomalies, up to 100\% for Poisson, under strong perturbations. In contrast, uniform noise appears more robust, as previously observed.
    
    \item Figures~\ref{linear anomaly no anomaly} and~\ref{missing anomaly no anomaly}: Linear trend perturbations and missing data perturbations result in relatively stable behavior across similar SNRs, indicating better robustness.
    
    \item Figure~\ref{amplitude anomaly no anomaly}: We observe that when changing the amplitude, the neural networks are more likely to misclassify an actual anomaly as a non-anomaly, rather than producing a false positive. While the neural networks seem to be very sensitive to this perturbation when considering class by class classification in Figure~\ref{amplitude}, we can see here that they are quite robust when considering only the detection of anomalies regardless of the urgency.
\end{itemize}

\section{Global Robustness}
\label{sec:global}

Besides local robustness, formal verification can be used more generally to help determine parts of the input domain where the classification algorithm is robust, therefore improving trustworthiness. Thus we worked on defining constraints for each anomaly histograms such that the classification is guaranteed. It means that for each class $x$, we want to define a finite set of constraints $(C_i^x)_{i=1}^L$ such that any histogram $h \in \mathbb{R}^M$ that satisfies all constraints $(C_i^x)_{i=1}^L$ is classified as $x$. Therefore, any time series or perturbed time series satisfying these constraints is guaranteed to be classified as $x$. 
Since we obtained results only for anomaly C, we restrict our analysis to this case.

To be as generic as possible, we want these constraints to be satisfied by a large majority of the histograms obtained from the training time series. That is the reason why, to define these constraints, we worked on generalizing some chosen characteristics of these histograms, like the minimum and maximum values possible at each point, the weighted sums, and local means when needed.

\begin{figure}[t]
\begin{subfigure}{.5\textwidth}
    \includegraphics[width=\textwidth]{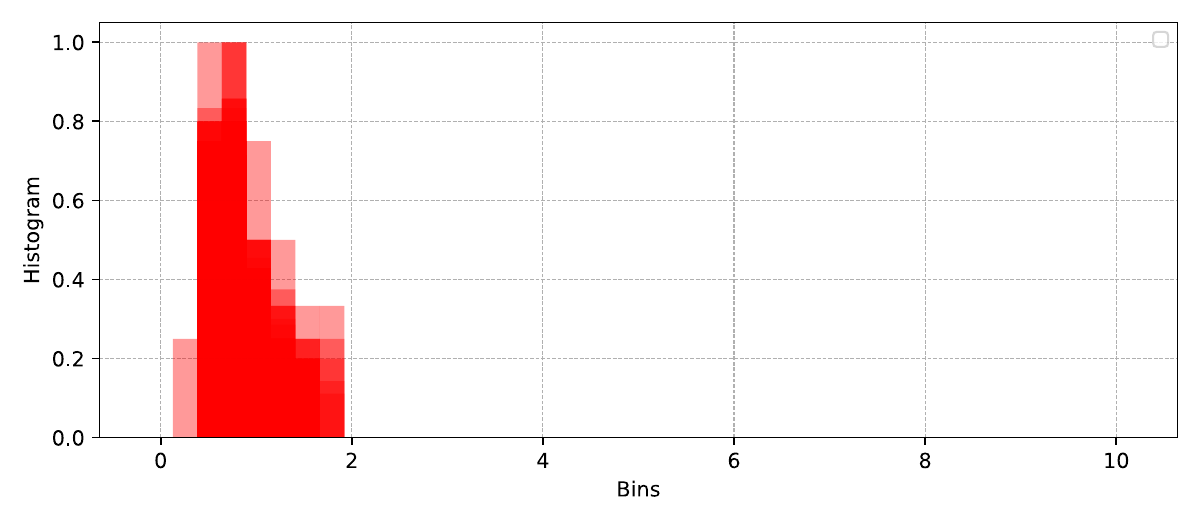}
    \caption{no C}
    \label{noC}
\end{subfigure}%
\begin{subfigure}{.5\textwidth}
\includegraphics[width=\textwidth]{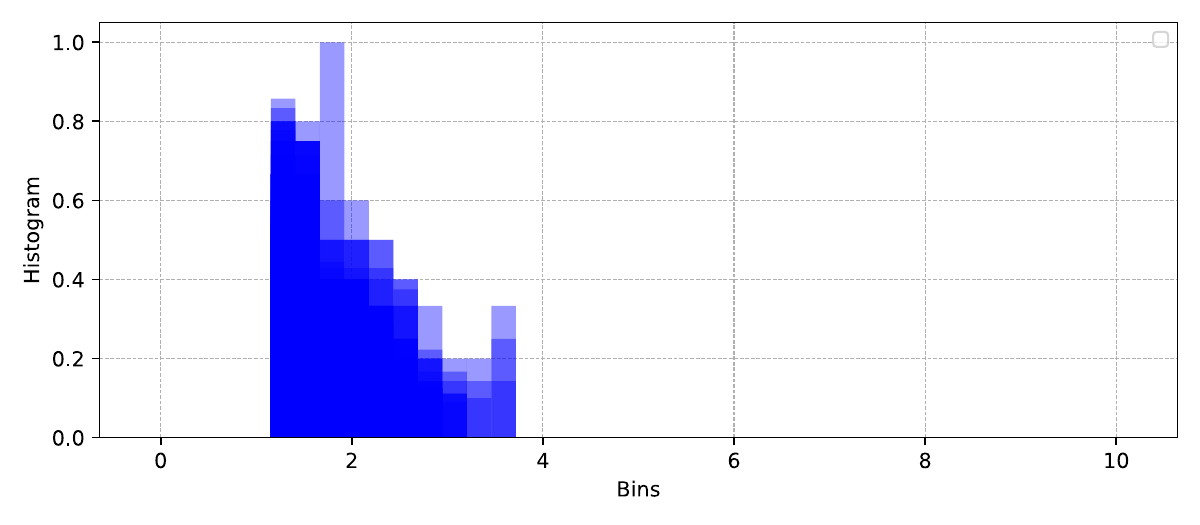}
    \caption{C1}
    \label{C1}
\end{subfigure}\\
\begin{subfigure}{.5\textwidth}
\includegraphics[width=\textwidth]{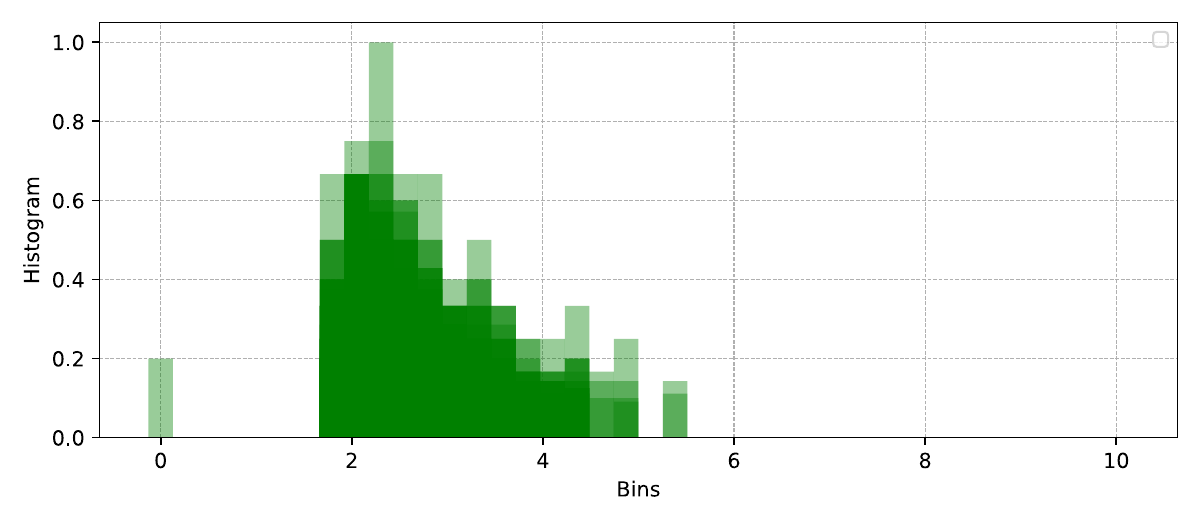}
    \caption{C2}
    \label{C2}
\end{subfigure}%
\begin{subfigure}{.5\textwidth}
\includegraphics[width=\textwidth]{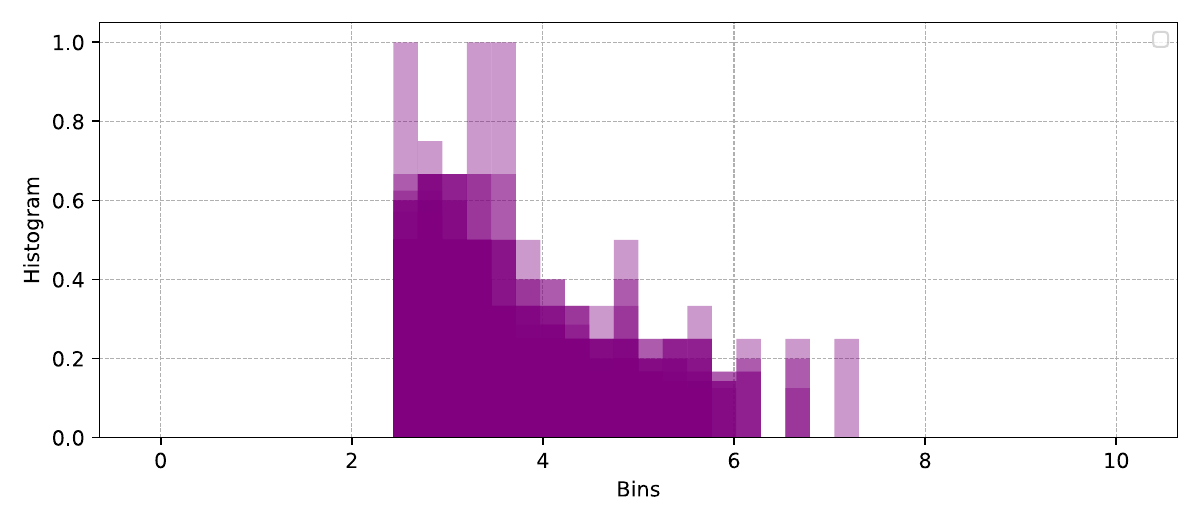}
    \caption{C3}
    \label{C3}
\end{subfigure}
\caption{All histograms of anomaly type C}
\label{all-peak-histo}
\end{figure}

First, if we look at all the anomaly C histograms obtained from the correctly classified time series of the dataset, displayed in Figure~\ref{all-peak-histo}, we can see that they have specific shapes and specific positions according to the class they belong to.

\begin{figure}
\begin{minipage}{0.49\textwidth}
    \centering
    \includegraphics[width=\linewidth]
    {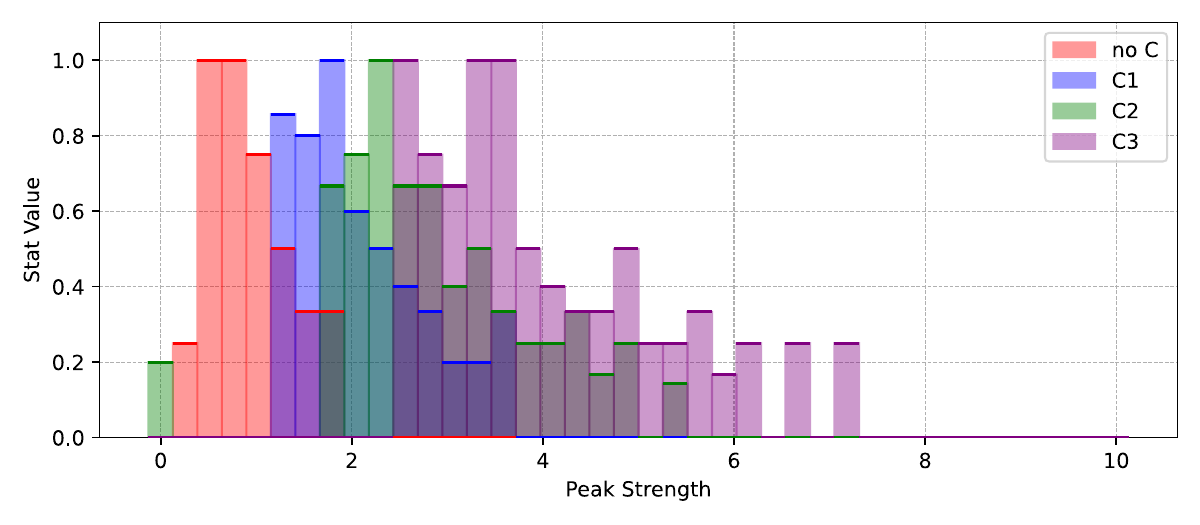}
    \caption{Upper bounds for classes of anomaly C}
    \label{upper-bounds-peak}
\end{minipage} \hfill
\begin{minipage}{0.49\textwidth}
    \centering
    \includegraphics[width=\linewidth]
    {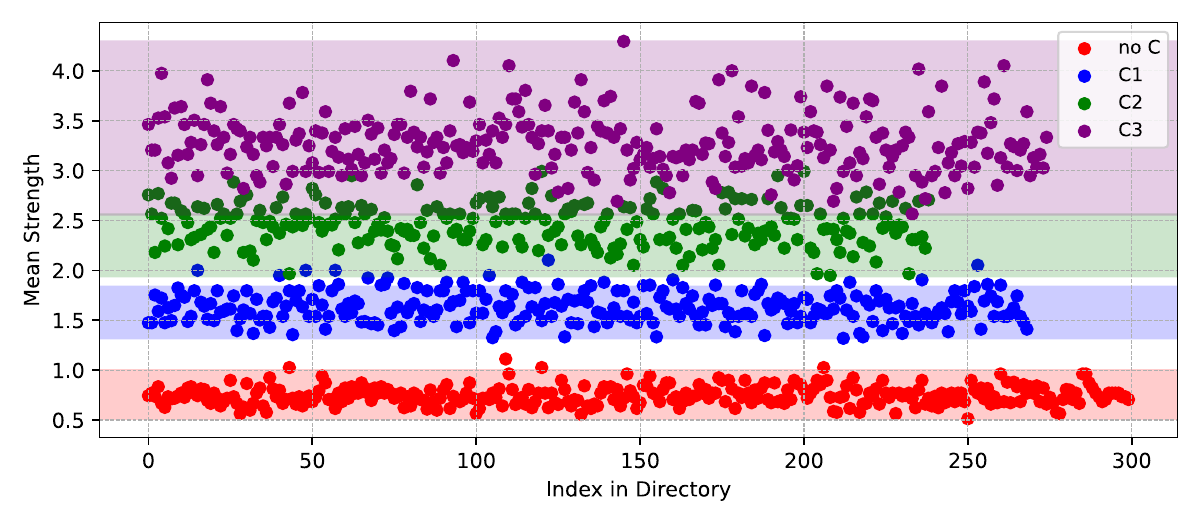}
    \caption{Weighted sums of all C histograms}
    \label{weighted-means-peaks}
\end{minipage}
\end{figure}

Based on these observations, we determined envelopes containing, for each class, all the histograms for that class. Figure~\ref{upper-bounds-peak} shows the upper bound at each point for each class of anomaly C, the lower bound always being 0.
This first constraint alone is not sufficient to capture any particular class, because the intersections between envelopes are too large. Further analysis revealed that in certain cases, for anomaly C, the classes could be distinguished using the weighted sum of the histogram. The weighted sum of a histogram $h \in {\mathbb R}^M$ is the product of bin values with their indices: $\sum_{i=1}^M h_i \times i$.
We analyzed the distribution of the weighted sums of all anomaly C histograms. Figure~\ref{weighted-means-peaks} shows that this feature does not clearly separate the different classes. However, we found that, when considered together with the envelopes defined above, well-chosen weighted sum intervals allow us to clearly discriminate between C1 and C3 anomalies, and also contribute to defining the `no C' case. These intervals do not necessarily include all the histograms of a class obtained from the dataset, but they were determined so as to exclude as few histograms as possible. In practice, only 4 histograms for `no C' (1.3\%) and 30 histograms for C1 (11\%) were excluded, while no histograms were excluded for C3.

\begin{figure}
\begin{minipage}[t]{0.49\textwidth}
    \centering
    \includegraphics[width=\linewidth]{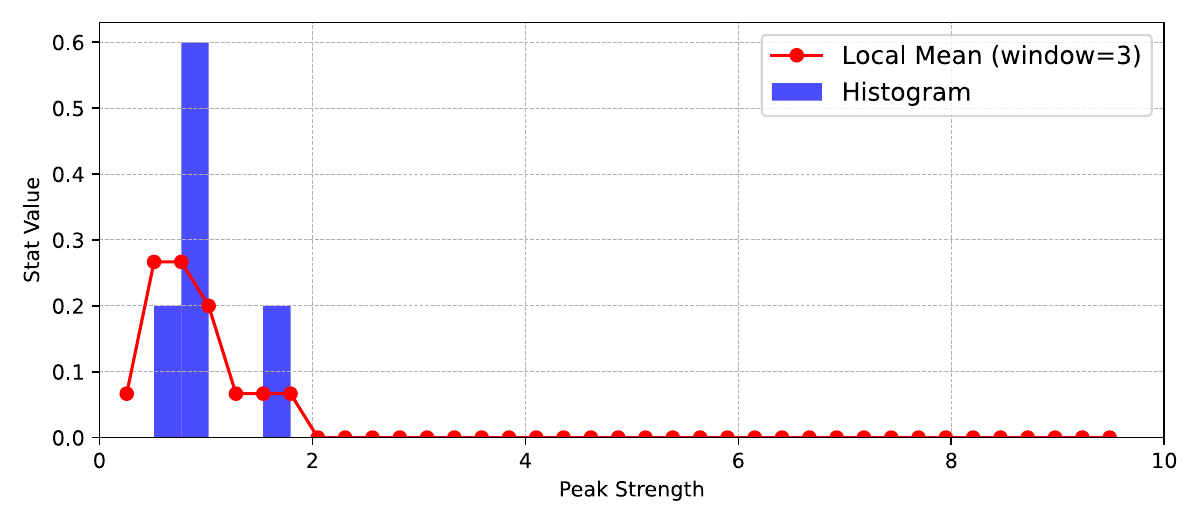}
    \caption{Sliding 3-sized window mean of a `no C' histogram}
    \label{example of window mean}
\end{minipage} \hfill
\begin{minipage}[t]{0.49\textwidth}
    \centering
    \includegraphics[width=\linewidth]{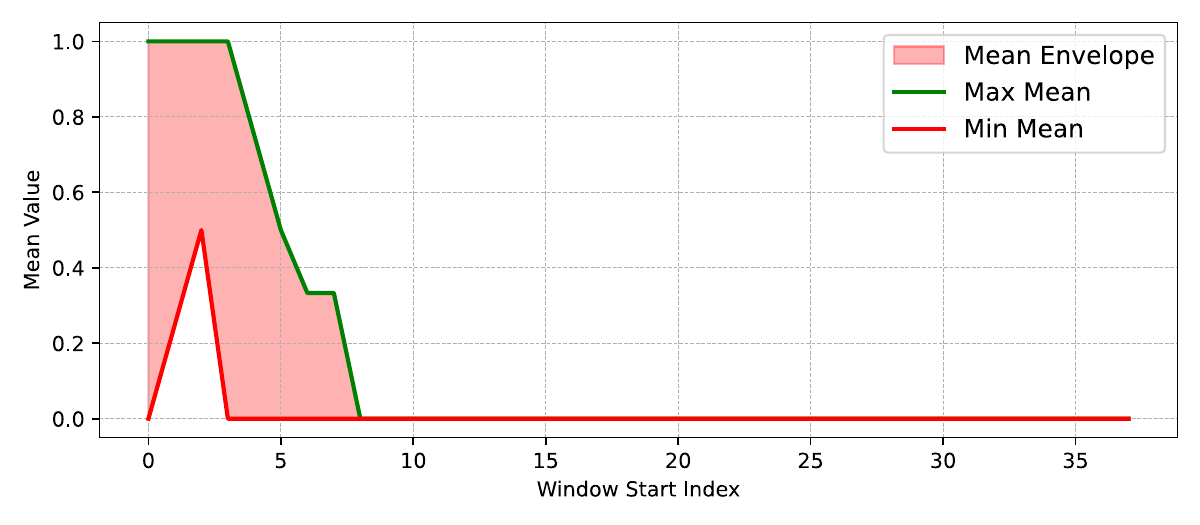}
    \caption{Minimum and maximum means for a sliding 3-sized window on all `no C' histograms}
    \label{sliding-window-no-C}
\end{minipage}
\end{figure}

For other classes, the weighted sum interval, even with the envelopes, is not sufficient, and a finer analysis at a local level had to be preferred. Specifically, for C2 and `no C' classes, we computed mean values over sliding windows of fixed sizes across the histogram. For a given $k \in \llbracket 1,M \rrbracket$, the mean over the sliding window of size $k$ is the vector $m \in {\mathbb R}^{M-k}$ defined by $\forall i \in \llbracket 1,M-k \rrbracket, m_i = \frac{1}{M}\sum_{j=i}^{i+M-1} h_j$.
Figure~\ref{example of window mean} shows the sliding 3-sized window mean of a `no C' histogram, and Figure~\ref{sliding-window-no-C} shows the results of minimum and maximum means for a sliding window of size 3 on all histograms of class `no C'.

Our experimentation shows that for class `no C', windows of size 3 and size 4 provide the best separation from the other classes, together with the envelope and the weighted sum constraint. Unfortunately, for class C2 as well as for all classes of anomaly D, no set of constraints has yet been found that ensures a robust classification.
To summarize, we formally proved the robustness of the neural network classifying anomaly C for the following cases.

\begin{center}
\begin{tabular}{|l|l|}
\hline
Class &	Constraints ensuring robustness \\
\hline
no C & `no C' envelope + \\
        & Weighted sum in $[0.51,1]$ +\\
        & `no C' envelope of 3-sized and 4-sized window means\\
\hline
C1	& C1 envelope +\\
        & Weighted sum in $[1.32,1.83]$\\
\hline
C3	& C3 envelope +\\
        & Weighted sum $[2.56,4.29]$\\
\hline
\end{tabular}
\end{center}

\section{Conclusion}

Our results demonstrate that using formal verification on a hybrid AI algorithm helps improve trustworthiness by providing guarantees essential to the deployment of critical fault detection systems in space applications.
One of the main challenges in applying formal methods to industrial AI systems lies in the data processing pipeline, which is often not amenable to formal reasoning. However, the design intuition behind these processes remains crucial in shaping a verification strategy. Our methodology enables a fine-grained analysis of the model’s behavior, identifying precisely where robustness breaks down and the system becomes unreliable, thus providing a practical measure of the model's robustness limits.

The class by class counterexamples produced by our verification tool have value in various contexts: for instance, when a certain classification's accuracy is critical, or when certain misclassifications can be tolerated. By grouping outputs into binary categories in our use case, we were able to draw meaningful local robustness conclusions, focusing on anomaly detection and not only on classification.

While the relevance of global robustness may be higher for functional specifications, our results suggest that this formal verification methodology can be extended to broader scenarios, helping define verified safe regions within the operational design domain (ODD).

Ultimately, our findings \edit{pave the way to the use of formal verification methods for AI in industrial contexts, even when the algorithms involving AI models are not fully designed with formal verification in mind, and can be seen as a first step to an end-to-end methodology compliant to recent standards on the development of systems involving AI components.}

\bibliographystyle{eptcs}
\bibliography{references}

\end{document}